\documentclass{article}
\usepackage{kr}

\usepackage{times}
\usepackage{soul}
\usepackage{url}
\usepackage[hidelinks]{hyperref}
\usepackage[utf8]{inputenc}
\usepackage[small]{caption}
\usepackage{graphicx}
\usepackage{amsmath}
\usepackage{amsthm}
\usepackage{booktabs}
\title{Consensus is Strategically Insufficient: Reasoning-Trace Disagreement as a Knowledge-Representation Signal}

\author{%
Micha{\l} Wawer$^{1}$ \and
Jaros{\l}aw A. Chudziak$^{1,2}$ \\
\affiliations
$^1$Laboratory of The New Ethos\\
Warsaw University of Technology, Warsaw, Poland\\
$^2$Institute of Computer Science\\
Faculty of Electronics and Information Technology\\
Warsaw University of Technology, Warsaw, Poland\\
\emails
\{michal.wawer.stud, jaroslaw.chudziak\}@pw.edu.pl
}

\begin{document}

\maketitle

\begin{abstract}
Multi-agent systems are commonly designed to reduce disagreement through voting, consensus protocols, debate, or fault-tolerant aggregation. We argue that this objective is insufficient for value-laden tasks, where disagreement may reflect genuine normative uncertainty rather than agent error. Building on prior work on reasoning-trace disagreement in human-AI collaborative moderation, we propose a knowledge-representation layer in which reasoning traces and agent decisions are abstracted into symbolic disagreement states. Given agents producing explicit reasoning traces and binary decisions, we distinguish four states according to reasoning similarity and conclusion agreement: convergent agreement, divergent agreement, convergent disagreement and divergent disagreement. These states support defeasible strategic routing rules. We instantiate the framework in content moderation and argue that disagreement-aware routing provides a bridge between sub-symbolic LLM deliberation and symbolic knowledge representation for multi-agent strategic reasoning.
\end{abstract}

\section{Introduction}

LLM-based multi-agent systems are increasingly used as collective reasoning architectures \cite{liang2024encouraging,chen2024reconcile} in which several agents deliberate, debate, or aggregate judgments before producing a final output \cite{kostka2025synergizinglogicalreasoningknowledge,sadowskiLegalReasoning2025}. Existing approaches typically treat inter-agent disagreement as a defect to be reduced through majority voting, additional debate rounds, or robust aggregation \cite{du2024improving,chen2024reconcile,liang2024encouraging,zheng2025rethinking,zhang2025multillmagents}. This is plausible for instrumental tasks where disagreement signals noise or reasoning failure. It is far less appropriate for value-laden tasks, where disagreement may be a stable property of the decision problem itself.
 
Content moderation is paradigmatic \cite{gajewska2026}. Decisions about harmful speech, group-directed language, or political criticism involve competing values, contextual interpretation, and socially situated judgment \cite{gorwa2020algorithmic,rieder2021fabrics,wang2022moderation,kuo2023unsung}. Annotator disagreement on such cases is not always error to be averaged away: it may reflect perspectival variation or genuine value pluralism \cite{kennedy2020constructing,sachdeva2022measuring}. The same observation applies to LLM agents: when differently profiled agents disagree, the disagreement may itself be informative.
 
We exploit this by extending our previous research \cite{wawer2026when} by introducing a knowledge-representation layer that abstracts agent reasoning traces and decisions into a small set of symbolic states and a defeasible policy that routes each state to a strategic meta-action. Three contributions follow. First, we reframe disagreement as a representable epistemic state of the multi-agent system rather than an aggregation obstacle \cite{austen1996information}. Second, we define a compact taxonomy along two dimensions: reasoning similarity and conclusion agreement, yielding four states: convergent agreement (\textit{CA}), divergent agreement (\textit{DA}), divergent disagreement (\textit{DD}), and convergent disagreement (\textit{CD}). Third, we associate these states with defeasible routing rules so the system reasons not only about what to decide, but about whether to decide, to inquire or to escalate.

\section{Disagreement as a Knowledge-Representation Signal}

We model an LLM-based multi-agent system \cite{wang2025mixtureofagents} as a finite set of agents $A=\{a_1,\dots,a_n\}$. For a case $c$ (a content item), each agent produces an output $O_i(c)=\langle r_i, d_i, v_i, \gamma_i\rangle$, where $r_i$ is an explicit reasoning trace, $d_i\in D$ is the agent's decision (here $D=\{\textsc{Keep},\textsc{Remove}\}$), $v_i$ is the value or perspective profile, and $\gamma_i$ is a confidence score. The KR layer treats $r_i$ as an observable justificatory artifact rather than a formal proof, in line with the standard view that agents have individual informational states while the system must determine a collective response \cite{wooldridge2009introduction,shoham2009multiagent}.
 
Two relations between agent outputs constitute the basic vocabulary. Let $sim(r_i,r_j)\in[0,1]$ denote the semantic similarity of two reasoning traces, with mean pairwise similarity $\overline{sim}(c)=\tfrac{2}{n(n-1)}\sum_{i<j}sim(r_i,r_j)$; given a threshold $\theta_s$, $HighSim(c)\equiv\overline{sim}(c)\geq\theta_s$ and $LowSim(c)\equiv\overline{sim}(c)<\theta_s$. The threshold is a policy parameter, not a universal semantic boundary. For conclusion agreement, let $p_d(c)=|\{a_i:d_i=d\}|/n$ and $p^*(c)=\max_{d\in D}p_d(c)$; given $\theta_a$, $Agree(c)\equiv p^*(c)\geq\theta_a$ and $Disagree(c)\equiv p^*(c)<\theta_a$. Conservative settings push $\theta_a$ toward unanimity; permissive settings accept supermajority.
 
Combining the two dimensions yields four symbolic states:
\begin{align*}
CA(c) &\equiv HighSim(c)\wedge Agree(c), \\
DA(c) &\equiv LowSim(c)\wedge Agree(c), \\
CD(c) &\equiv HighSim(c)\wedge Disagree(c), \\
DD(c) &\equiv LowSim(c)\wedge Disagree(c).
\end{align*}
 
\begin{figure}[t]
\centering
\includegraphics[width=0.9\linewidth]{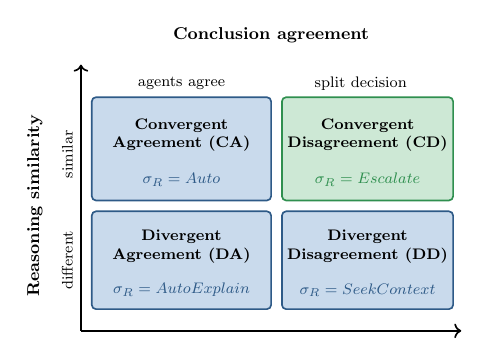}
\caption{The four disagreement states arise from combining reasoning similarity with conclusion agreement. Each state has a default meta-action $\sigma_R$ (Sec.~\ref{sec:rules}). Convergent disagreement is treated as the strongest candidate signal of value-laden conflict.}
\label{fig:state-space}
\end{figure}

These are not merely empirical clusters, they are symbolic abstractions of the multi-agent system's epistemic situation, available to a controller. As in formal argumentation and nonmonotonic reasoning, conflicting reasons may support different conclusions. We treat the resulting structure as a representable object \cite{dung1995acceptability,rahwan2009argumentation,brewka1997nonmonotonic}. The state of greatest interest is $CD(c)$: when agents reason similarly but conclude differently, the residual disagreement is unlikely to be a difference of interpretation. It more plausibly reflects different value weightings on a shared description of the case, a candidate signature of normative pluralism rather than error. By contrast, $DD(c)$ suggests ambiguity or unstable interpretation; $DA(c)$ suggests robustness through independent reasons; $CA(c)$ is the most straightforward case for automatic resolution. Figure~\ref{fig:state-space} summarizes the taxonomy together with the default meta-actions defined next.
 
\section{Defeasible Strategic Routing Rules}
\label{sec:rules}

The disagreement states do not by themselves determine the moderation label. They determine a \emph{meta-action}: the system reasons about whether to commit to an automatic decision at all. Let $d^*(c)=\arg\max_{d\in D}p_d(c)$ denote the most-supported decision. We consider four meta-actions: $Auto(c,d^*)$, automatically accept the strongest decision; $AutoExplain(c,d^*)$, accept the strongest decision but preserve diverse explanations; $SeekContext(c)$, request additional information or another deliberation round; and $Escalate(c)$, forward the case to human judgment.
 
We use $\Rightarrow$ to denote a default inference whose consequent normally holds but may be overridden by stronger policy or risk constraints, in the spirit of nonmonotonic reasoning \cite{brewka1997nonmonotonic}. The base routing policy is:
\begin{align}
R_1: &\quad CA(c) \Rightarrow Auto(c,d^*), \label{eq:r1}\\
R_2: &\quad DA(c) \Rightarrow AutoExplain(c,d^*), \label{eq:r2}\\
R_3: &\quad DD(c) \Rightarrow SeekContext(c), \label{eq:r3}\\
R_4: &\quad CD(c) \Rightarrow Escalate(c). \label{eq:r4}
\end{align}
 
Rule $R_1$ captures the easy case: justificatory and decisional convergence jointly justify automation. $R_2$ handles agreement-through-diverse-reasons; because the reasons differ, the system preserves explanation diversity rather than collapsing them into a single rationale, which matters when different stakeholders require different explanations \cite{amgoud2009using,rahwan2009argumentation}. $R_3$ handles divergent disagreement: the system may not yet have a stable representation of the case, so context acquisition typically dominates immediate escalation. $R_4$ is the central rule. In $CD$, agents share a broadly similar interpretation but transform it into different decisions; forcing consensus \cite{denisovblanch2025consensus} here may conceal rather than resolve a normative conflict.
 
Defeasibility is essential. Even $CA(c)$ may be overridden when content is legally sensitive or predicted harm is high; conversely, $CD(c)$ may not warrant escalation in low-risk cases with high escalation cost: $HighRisk(c)\Rightarrow Escalate(c)$, $LegalRequirement(c)\Rightarrow Escalate(c)$, and $LowRisk(c)\wedge HighEscCost(c)\Rightarrow AutoExplain(c,d^*)$. The final meta-action results from interaction between disagreement-state rules and domain rules, in line with classical defeasible-reasoning architectures \cite{brewka1997nonmonotonic,dung1995acceptability}. Decision-theoretically, each meta-action has a different cost profile automation risks illegitimate decisions, $SeekContext$ adds latency, $Escalate$ consumes scarce institutional capacity and the disagreement state provides a structured signal for allocating these costs, complementing judgment-aggregation perspectives that combine votes but do not, by themselves, decide \emph{whether} to aggregate \cite{grossi2014judgment}.

\begin{figure*}[t]
\centering
\includegraphics[width=0.9\linewidth]{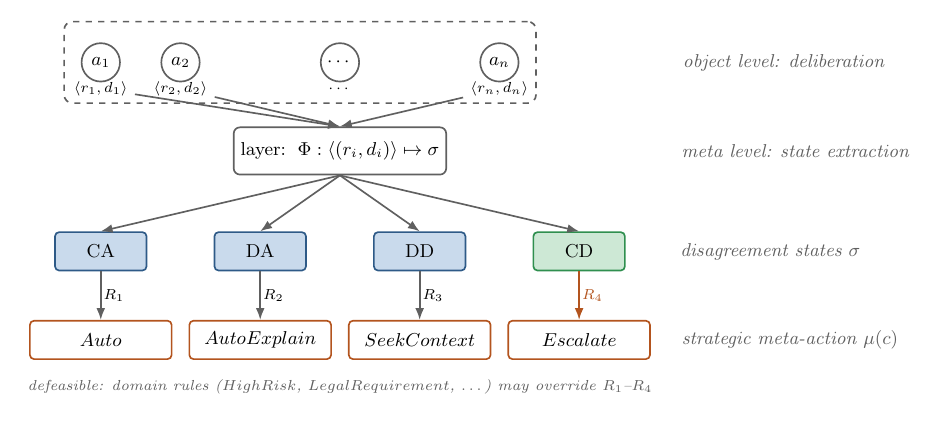}
\caption{Architecture of the disagreement-aware controller. LLM agents deliberate at the object level, producing reasoning traces and decisions $\langle r_i,d_i\rangle$. The KR layer applies the abstraction $\Phi$ to extract one of four symbolic states $\sigma\in\{CA,DA,DD,CD\}$. Defeasible rules $R_1$--$R_4$ then map each state to a strategic meta-action; the convergent-disagreement path (highlighted) defaults to $Escalate$, but any rule may be overridden by domain-level defaults.}
\label{fig:pipeline}
\end{figure*}

\section{Empirical Faithfulness Check: Content Moderation}
\label{sec:setup}

The framework above is normative, it prescribes how a controller should react to disagreement structure. We need to check whether the symbolic abstraction is faithful in a weaker but useful sense: do the four states track empirically distinct epistemic situations, in particular ones that humans also find different? This is a sanity check on the KR layer, not a benchmark of the routing policy.
 
We reuse the experimental setup of \cite{wawer2026when}. For each content item $c$, five LLM agents are instantiated from the same base model and differentiated by system prompts encoding distinct moderation perspectives: \emph{harm-focused}, \emph{context-sensitive}, \emph{community-norms}, \emph{free-expression}, and \emph{legal-framework}. This isolates value-profile differences from base-capability differences. Each agent produces $\langle r_i,d_i,v_i,\gamma_i\rangle$ where $d_i\in\{\textsc{Keep},\textsc{Remove}\}$ and $r_i$ contains the agent's interpretation, considerations, value trade-offs and conclusion.
 
We use the Measuring Hate Speech corpus \cite{kennedy2020constructing,sachdeva2022measuring}, which preserves annotator variation and supports perspectivist analysis. We sample $n=600$ items stratified by human annotator disagreement. Reasoning traces are embedded into a shared vector space, and pairwise cosine similarity yields $\overline{sim}(c)$; the decision distribution yields $p^*(c)$. Each case is labeled with one of the four symbolic states, giving the abstraction
\[
\Phi:\langle (r_i,d_i)_{i=1}^n\rangle \;\longmapsto\; \sigma\in\{CA,DA,DD,CD\}.
\]
The faithfulness check asks two questions about $\Phi$: (i) do cases assigned to different states differ in human disagreement, and (ii) does the structural distinction $\Phi$ provide information beyond a magnitude-only baseline that ignores conclusion structure?

\section{Preliminary Results and Evaluation}
\label{sec:results}

Table~\ref{tab:taxonomy} reports the distribution of cases across states and the corresponding mean human annotator disagreement $\bar{d}$. The ordering predicted on conceptual grounds: $DA<CA<DD<CD$ is preserved: divergent agreement is the most stable, convergent disagreement the least. The two disagreement states $\{CD,DD\}$ are jointly separated from the two agreement states $\{CA,DA\}$ with effect size Cohen's $d=0.80$ ($p<10^{-11}$, $n=600$), suggesting that the structural abstraction tracks something humans also pick up on.
 
\begin{table}[t]
\centering
\small
\begin{tabular}{lccc}
\toprule
\textbf{State} & \textbf{Description} & \textbf{$n$} & \textbf{Mean $\bar{d}$} \\
\midrule
$DA(c)$ & Divergent agreement      & 118 & 0.351 \\
$CA(c)$ & Convergent agreement     & 24  & 0.638 \\
$DD(c)$ & Divergent disagreement   & 382 & 0.751 \\
$CD(c)$ & Convergent disagreement  & 76  & 0.782 \\
\bottomrule
\end{tabular}
\caption{Distribution of cases over symbolic states and mean human annotator disagreement $\bar{d}\in[0,1]$. Predicted ordering $DA<CA<DD<CD$ is preserved.} 
\label{tab:taxonomy}
\end{table}
 
\begin{figure}[t]
\centering
\includegraphics[width=0.95\linewidth]{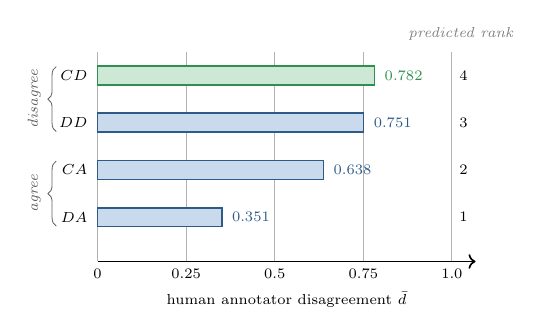}
\caption{Observed mean human disagreement $\bar{d}$ per symbolic state, with the conceptually predicted rank order on the right ($1{=}$lowest, $4{=}$highest). The qualitative ordering $DA<CA<DD<CD$ is preserved; 
}
\label{fig:predicted-observed}
\end{figure}

A natural baseline is to use only the magnitude of disagreement, e.g., $1-\overline{sim}(c)$, ignoring conclusion structure. Table~\ref{tab:routing} compares the two as predictors of high human disagreement: category-based routing achieves higher F1 than divergence-only and substantially exceeds chance. Divergence-only achieves high recall but lower precision: it flags many cases where agents reason differently without that necessarily corresponding to human disagreement. This is precisely what the $CD/DD$ distinction captures a purely metric account loses the second axis (whether agents nevertheless converge on a decision), which is exactly the dimension that separates likely-normative cases ($CD$) from likely-ambiguous ones ($DD$). Figure~\ref{fig:predicted-observed} visualizes the qualitative ordering.
 
\begin{table}[t]
\centering
\small
\begin{tabular}{lccc}
\toprule
\textbf{Predictor} & \textbf{Precision} & \textbf{Recall} & \textbf{F1} \\
\midrule
Category-based escalation & 0.401 & 0.845 & \textbf{0.548} \\
Divergence only           & 0.347 & 0.915 & 0.503 \\
Random baseline           & 0.333 & 0.505 & 0.401 \\
\bottomrule
\end{tabular}
\caption{Flagging high human-disagreement cases. Category-based routing uses $\Phi$; divergence-only uses $1-\overline{sim}(c)$ thresholded at the same operating point.} 
\label{tab:routing}
\end{table}
 
The check is preliminary, a single corpus, prompt-based agent differentiation and embedding-based similarity. Stronger conclusions require independently parameterized agents and alternative similarity functions. The result we treat as load-bearing is qualitative the ordering and the agreement/disagreement gap, rather than the specific F1 numbers.
 
\section{Discussion: From Consensus to Strategic Escalation}
\label{sec:discussion}

The framework reframes the design goal of LLM-based multi-agent systems. A consensus-seeking system asks how agents can be made to agree; a disagreement-aware system asks what the structure of disagreement implies about the appropriate next action. Reasoning traces are central to this shift: a vote alone does not reveal whether agents disagree because they misread the case or weigh shared considerations differently. By comparing traces and decisions jointly, the controller distinguishes interpretive from evaluative disagreement in a manner reminiscent of argumentation frameworks where conclusions depend on the structure of supporting and attacking reasons \cite{dung1995acceptability,rahwan2009argumentation,amgoud2009using}.
 
The state $CD(c)$ does the most strategic work. In factual tasks it would look like inconsistency, in normative tasks it more plausibly indicates that agents share a description of the case and differ in value prioritization. Collapsing such cases into a single automatic decision risks hiding legitimately contested situations. Escalation here is not a failure of automation but a rational meta-action under normative uncertainty. Symmetrically, $DD(c)$ should typically trigger context acquisition rather than human review, because the system has not yet stabilized a representation. This $CD/DD$ asymmetry is a chief benefit of the taxonomy: both involve disagreement, but call for different strategies.
We do not claim LLM reasoning traces are formal proofs, nor that semantic similarity captures logical equivalence. The KR layer is a pragmatic interface between sub-symbolic deliberation and symbolic strategic control \cite{grossi2014judgment,brewka1997nonmonotonic}.

Multi-agent debate, round-table consensus and Byzantine-tolerant aggregation \cite{du2024improving,liang2024encouraging,chen2024reconcile,zheng2025rethinking} share a design assumption that disagreement is a transient state to be resolved before output. The empirical line on \emph{perspectivist NLP and content moderation} \cite{kennedy2020constructing,sachdeva2022measuring,gorwa2020algorithmic} challenges this at the data level, treating annotator disagreement as informative rather than noisy. We transfer this perspectivist stance from data to system architecture, making disagreement a representable state of the multi-agent system rather than a defect.
Judgment aggregation \cite{list2002aggregation,grossi2014judgment} combines individual judgments without exposing the reasoning behind each. Formal argumentation \cite{dung1995acceptability,rahwan2009argumentation,amgoud2009using} exposes that reasoning as attack and support relations. Our four-state projection sits between the two: it preserves enough reasoning structure to distinguish shared from divergent interpretations (beyond what aggregation sees), but treats interpretations as propositional abstractions rather than full argumentation graphs. 

Several limitations are worth flagging. Prompt-based perspective differentiation may underrepresent the heterogeneity of independent agents. Embedding similarity is a coarse proxy for reasoning equivalence. The routing rules are hand-designed defaults rather than learned or formally verified policies. The empirical check covers a single domain. Promising directions follow from each. The KR layer can be enriched with explicit beliefs, preferences, and norms. Reasoning traces can be coupled to argumentation graphs, so support, attack, and undercutting are detected directly. Explicit cost models would let escalation choices be analyzed game-theoretically. Finally, the faithfulness check should be replicated in domains such as medical triage and legal assistance, where disagreement plausibly carries similar structural significance.

\section{Conclusion}

LLM-based multi-agent systems are typically designed to suppress disagreement. This goal is strategically insufficient in value-laden tasks, where disagreement may be a stable property of the case rather than a transient defect. We proposed a knowledge-representation layer that abstracts agent reasoning traces and decisions into four symbolic states:$CA$, $DA$, $DD$, $CD$ - and a defeasible policy mapping each to a strategic meta-action.
 
The result explicit interface between sub-symbolic LLM deliberation and symbolic strategic control: the system reasons not only about what to decide, but about when to decide, when to inquire, and when to escalate. A faithfulness check in content moderation is consistent with the claim that the structure of disagreement carries information its magnitude does not, with convergent disagreement most strongly tracking human normative conflict. Natural extensions include coupling traces to argumentation graphs, learning the routing rules and experimenting with different LLMs.

\bibliographystyle{kr}
\bibliography{references}

\end{document}